# Convolutional neural network based on sparse graph attention mechanism for MRI super-resolution

Xin Hua[a] Zhijiang Du[a] Hongjian Yu[a] Jixin Ma[a]

a State Key Laboratory of Robotics Technology and Systems, Harbin Institute of Technology, Harbin 15000, Heilongjiang, China

**Abstract**

  Magnetic resonance imaging (MRI) is a valuable clinical tool for displaying anatomical structures and aiding in accurate diagnosis. Medical image super-resolution (SR) reconstruction using deep learning techniques can enhance lesion analysis and assist doctors in improving diagnostic efficiency and accuracy. However, existing deep learning-based SR methods predominantly rely on convolutional neural networks (CNNs), which inherently limit the expressive capabilities of these models and therefore make it challenging to discover potential relationships between different image features. To overcome this limitation, we propose an A-network that utilizes multiple convolution operator feature extraction modules (MCO) for extracting image features using multiple convolution operators. These extracted features are passed through multiple sets of cross-feature extraction modules (MSC) to highlight key features through inter-channel feature interactions, enabling subsequent feature learning. An attention-based sparse graph neural network module is incorporated to establish relationships between pixel features, learning which adjacent pixels have the greatest impact on determining the features to be filled. To evaluate our model's effectiveness, we conducted experiments using different models on data generated from multiple datasets with different degradation multiples, and the experimental results show that our method is a significant improvement over the current state-of-the-art methods.

**Key Word**：MRI、 super-resolution reconstruction、Graph Neural Network、Sparse graph

## 1.Introduction

  Medical imaging technology plays a critical role in modern medical systems, with magnetic resonance imaging (MRI) being widely used in clinical medicine due to its safety and comprehensive information. High-resolution MRI images can provide detailed structural information that aids in subsequent medical diagnoses [1-2]. However, obtaining HR-MRI remains challenging due to equipment limitations and scanning time. Currently, there are two general approaches to improve MRI resolution: using high-performance equipment or enhancing the scanning method, which can increase costs. Alternatively, super-resolution reconstruction techniques can obtain HR images from given LR images, overcoming equipment limitations and meeting clinical requirements at a lower cost and with greater efficacy. While many traditional methods, such as interpolation [3-4], dictionary-based [5-6], and patch-based super-resolution [7-8], have been applied in brain MRI to obtain super-resolution results, the magnified images produced by these methods have smoothness and blurring issues, leading to the loss of detailed anatomical information. In recent years, deep learning has rapidly developed, leading to its application in image super-resolution by numerous researchers. Excellent results have been achieved with models such as DLRRN [9], DRLM[10], FA-GAN [11], CSNLN [12], and CSSFN [13]. These models exhibit incredible performance in super-resolution reconstruction.

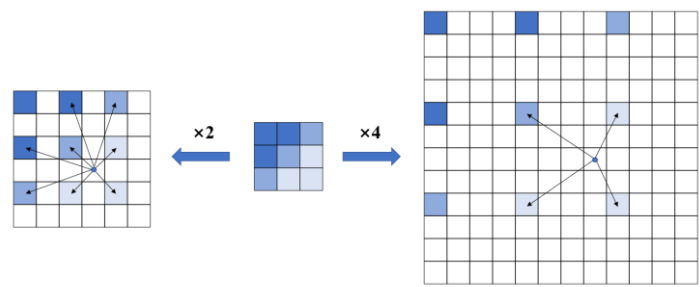

Fig1 Relationship between pixels after image scaling

  The current approach utilizes existing pixel data to infer the value of a blank pixel, while considering the proximity of the blank pixel to existing ones. The closer the blank pixel is to an existing pixel, the greater its influence. To fill a blank pixel, we take into account the existing pixel values within a certain range as a reference for location and value, while decreasing the number of values considered as image magnification increases.

Graph Neural Networks (GNN) are deep learning models capable of processing graph data[14]. Unlike traditional models, GNNs can capture both local and global information of nodes and edges by aggregating information of nodes and edges to better describe the structure and characteristics of graphs. This allows GNNs to handle non-Euclidean spatial data effectively. As a result, GNNs have the ability to handle changing graph structures, such as dynamic social networks and chemical molecules, in various fields[15-16]. Recently, researchers have applied GNNs to image processing[17-18], where local and global hierarchies are established using CNN models. These models extract features from local data, which are then used as nodes for constructing GNN models. Global feature learning is performed as the entire graph is constructed in this manner.The contributions of this work can be summarized as follows:

**(1)** We propose a variety of convolution kernel feature extraction module, through different sizes of convolution kernels and the design of using different kinds of operators for feature extraction, through different sizes of convolution kernels to obtain different field of view range, different kinds of operators are conducive to the extraction of features with unique properties in the features, through the residual structure, so that the combination of different feature extraction content, so that the final feature results more rich information content.

**(2)** We design a Multiple sets of cross-feature extraction modules to group the features in different channels, perform softmax and intelligent point formation between the initial two groups, and accumulate the results into the next group, through multiple accumulation, the key features in the image will be highlighted to facilitate subsequent feature learning.

**(3)** We design an Information Separation Hybrid Attention Module integrating Nonlinear Scale Normalization networks and Graph Neural Networks (GNN). By incorporating both global and local attentional mechanisms into one unified framework, we aim to simultaneously learn global relationships among elements across layers and exploit local connections to generate superior predictions.

**(4)** In order to validate the effectiveness of our newly introduced model, we carried out a rigorous series of ablation tests. The experimental results showed that our model consistently outperforms existing image SR techniques in terms of both Peak Signal-To-Noise Ratio (PSNR) and Structure Similarity Index Measure (SSIM), indicating its superior ability to reconstruct higher quality images.

## 2. Related work

### 2.1 Convolutional neural networks for super-resolution applications

Many CNN-based networks have been proposed to facilitate medical image super-resolution reconstruction research.BSRN [19] consists of four parts: shallow feature extraction, deep feature extraction, multi-layer feature fusion and reconstruction. Texture and feature information at different levels of abstraction are extracted from the feature maps of the convolutional layers at different stages, and these feature information is finely utilized through fusion operations to recover higher quality SR images.BSRN improves on the local module based on the network structure of RFDN [20], which enhances the reconstruction capability of the model while compressing the number of parameters and computational effort of the model.Sparse Mask SR (SMSR)[23]achieves pruning redundancy computationally by learning sparse masks. The sparse mask discriminates the significant regions in the feature map while the channel mask learns the redundant channels (i.e., unimportant regions). The model is precisely located and skipped through the redundant computational null fields, achieving a reduction in computation while maintaining model performance. Zhao [27] proposed a new channel segmentation and serial fusion network (CSSFN) for super-resolution of single MR images. The proposed CSSFN divides the hierarchical features into a series of sub-features, which are then integrated together in a serial manner. Dense global feature fusion (DGFF) in the model is used to integrate the intermediate features, which further facilitates the information flow transfer in the network and helps to improve the efficiency of the network. The RFANet[21]network combines multiple residual modules together and propagates features on each local residual branch by adding jump connections. An enhanced spatial attention block is also proposed, which adaptively rescales the elements according to the spatial context in order to make the residual features more focused on the critical spatial content. The EBPN model[22]establishes two branches of embedded residual modules, which are used to recover low and high frequency information, respectively. The embedded residual modules are used to pass the information that is difficult to recover in the previous layer into the deeper layers for recovery. EFDN (edge-oriented feature distillation network) [29] improves on RFDN by improving efficiency using ECB (edge-oriented convolution block) instead of SRB (shallow residual block) and using a reparameterization technique to improve SR performance.

### 2.2 Attentional mechanisms applied in super-resolution

In order to better capture the features that have a key impact on image reconstruction such as high-frequency features or similar features in the learned images, and to improve the image reconstruction effect by designing the corresponding attention mechanism for learning, many researchers have done a lot of work in designing the attention mechanism. Yiqun Mei et al. proposed an attention mechanism, Cross-Scale Non-Local Attention[24], which combines local features, within-scale non-local features and cross-scale non-local features to build a new attention module to find more a priori information in images. The model uses Information multi-distillation block for feature extraction module with overall residual structure, each part of the model extracts a part of useful features, and the rest of the features continue to be extracted downward through convolution. aware channel attention, which adjusts the weight of each channel by the sum of the standard deviation and mean of the channel to integrate the overall feature extraction, and finally adjusts the number of channels by using the 1x1 convolutional dimensionality reduction[25]. Yulun Zhang[26] designs a Context Reasoning Attention in which the attention mechanism uses graph neural networks to construct pixel-image connections. In this case the convolution kernel is adaptively adjusted according to the global features. The NJUST_ESR[28] project employs a vision transformer (ViT) that takes advantage of both convolution and multi-headed self-attention. To extract both local and global information, the project utilizes a hybrid module consisting of a ViT block and an inverse residual module, which is stacked multiple times to learn features. Additionally, Bin Xia et al[32] introduced a new method called Efficient Non-Local Contrast Attention (ENLCA), which offers remote vision modeling and captures more relevant non-local features. ENLCA consists of, Efficient Non-Local Attention (ENLA) and Sparse Aggregation. For sparse aggregation, the input features are multiplied by an amplification factor to focus on informative features, resulting in an exponential growth of the approximate variance. Thus features can be applied to separate relevant and irrelevant features and improve learning efficiency.

## 2.3 Graph Neural Networks

Due to the good relational reasoning ability of graph networks, many graph-based methods have been proposed for super-resolution image reconstruction studies Squeezed and Stimulated Attention Network (SERAN)[30] uses a second-order attention pooling operation to compress global spatial information into a global description, which allows the model to focus on the more informative regions and structures in the image. Attention is inferred by establishing relationships between primitives (primitives) and applying a graph convolutional network (GCN) to reason and obtain primitive relationships. Wang et al[33] proposed a method for representing node features as vector representations of nodes using a combination of node features and graph topological information is proposed, and an attention mechanism is used to learn the magnitude of the weights between individual feature channels. Yao et al[34]. used graph convolution to build a multi-feature attention module, which contains three parts: neighborhood attention, edge attention, and neighborhood-edge attention. Neighborhood attention is used to reflect the importance coefficients of the neighbors of the centroids. Edge attention, on the other hand, is used to learn the importance coefficients of edges to centroids. Neighbor-edge attention is mainly used for feature fusion, i.e., presenting the features of the central node. Combining the three components effectively extracts local spatial features. Zhou et al.[35] build a cross-scale internal graph neural network (IGNN). A cross-scale graph is dynamically constructed, and the cross-scale graph is constructed by using k-nearest neighboring in downsampling to query patches to search for k nearest neighboring patches for relationship building and aggregating them according to the adaptive way. and aggregate them adaptively according to the edge labels of the constructed graph. In this way, the low-resolution patches can obtain information about the adjacent high-resolution patches to help them in information recovery. Liu et al[36] proposed a dual learning based graph neural network, where the network uses patches in an image as graph neural network nodes, learns by aggregating feature patches that are similar across scales, and uses a dual learning strategy to refine the reconstruction results. Bao et al.[37] designed an end-to-end attention-driven graph neural network with a dynamic graph block in the model to establish the cross-scale relationships of patches in different regions. The spatially non-local self-similarity information is explored using the spatially aware dynamic graph unit in the channel attention and spatial dynamic graph block.

## 3. Proposed method

As shown in Figure 2, the whole network consists of shallow feature extraction, deep feature extraction, and image reconstruction. This architecture design has been widely used in previous works [54-55].In the Shallow feature extraction module,the network first inputs a low-resolution image $I_{LR} \in R^{3 \times H \times W}$, where H and W represent the low-resolution image height and width, respectively. The low-resolution image is fed into a shallow content feature extraction module, denoted by $F_{SFCE}(\cdot)$, which consists of a 3x3 convolution for initial feature extraction of the image.

$$T_0 = F_{SFCE}(I_{LR}) \tag{1}$$

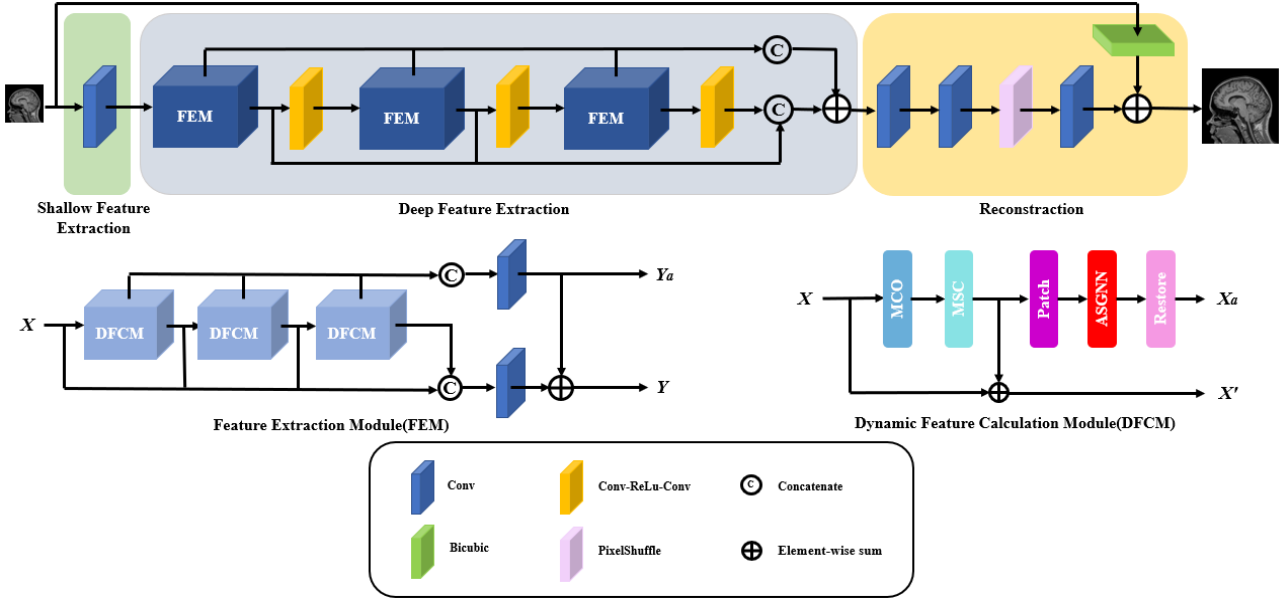

Fig2　The overall architecture of A and the structure of FEM and DFCM

The deep feature content extraction module consists of three Feature Extraction Modules (FEM) for deep content extraction. The transition between FEM is accomplished using conv-relu-conv, which serves the purpose of feature learning and also changes the number of feature channels. We have chosen not to include pooling operations in the model because we believe that doing so would result in the loss of image features, which cannot be fully recovered in the subsequent image reconstruction. Each FEM in the model contains three Dynamic Feature Calculation Modules (DFCM) for feature extraction. The DFCM further perform feature extraction through multiple convolution operator feature extraction modules and multiple sets of cross-feature extraction modules. To prevent feature loss as the depth of the model increases, we use the residual network for chunking the extracted features before passing them into the Attention-based sparse graph neural network module (ASGNN) to establish the connection between global and local features. Then, as shown in Equation 2 and Equation 3 we extract the features $X_a$ filtered by the attention mechanism, merge X and $X_a$ in each DFCM in the FEM, and convolve the merged features while performing intelligent summation to transfer the features Y and $Y_a$.

$$X_a = F_{\text{Restore}}(F_{ASGNN}(F_{Patch}(F_{MSC}(F_{MCO}(X)))))$$
$$X' = X + F_{MSC}(F_{MCO}(X)) \tag{2}$$

Similarly, the Y and $Y_a$ from each FEM are merged and intelligently summed in the network before passing the resulting features into the recovery section. In summary, we have omitted pooling operations from our model to avoid losing image features, and we have utilized a combination of convolution operators, cross-feature extraction modules, residual networks, and attention mechanisms to perform effective deep content extraction.

$$Y_a = Conv(Concat(X_{a1}, X_{a2}, X_{a3}))$$
$$Y = Conv(Concat(X, X'_1, X'_2, X'_3)) + Y_a \tag{3}$$

As shown in Equation 4, $Y_{a1}$, $Y_{a2}$ and $Y_{a3}$ from each FEM are combined to obtain $I_{att}$, and the combined features $I$ from each FEM such as $Y_1$, $Y_2$ and $Y_3$ are summed to obtain all features $I_{DE}$.

$$I_{att} = Concat(Y_{a1} + Y_{a2} + Y_{a3})$$
$$I = Concat(Y_1, Y_2, Y_3, Conv(\text{Re}Lu(Conv(Y_3))))$$
$$I_{DE} = I_{att} + I \tag{4}$$

In the Restoration module, $I_{DE}$ is sent to the feature recovery reconstruction module to complete the super-resolution reconstruction of the image. This process can be described as:

$$I_{SR} = F_{RE}(I_{DE}) + F_{bicubic}(I_{LR}) \tag{5}$$

where $F_{RE}(\cdot)$ denotes the recovery module, which consists of a pixel shuffle operation and two 3 × 3 convolutions, adjusted using the feature content of the 3 × 3 convolutions. Where $F_{bicubic}(\cdot)$ denotes the dual triple interpolation upsampling operation. By adding the result after $I_{LR}$ double triple interpolation upsampling to the final feature recovery result, the image reconstruction performance of the network is improved.

## 3.1 Multiple convolution operator feature extraction modules

After the initial extraction of features through 3×3 convolutional layers, 1×1 convolutions are utilized to adjust the number of feature channels, ensuring consistent channel sizes for subsequent convolutional kernels. Multiple types of kernels, including 3×3 convolutional kernels, 5×5 convolutional kernels, Laplace convolutional kernels, and edge convolutional kernels, are then applied to the features in order to extract different characteristics. Inspired by SPSR[56], the formula for the edge convolutional kernels is illustrated in Figure 2, with separate image edge feature extractions performed in the X and Y axes. To enhance the accuracy of feature edge extraction, left and right tilt operators are added.

$$G_x = \begin{bmatrix} 0 & 0 & 0 \\ 1 & 0 & -1 \\ 0 & 0 & 0 \end{bmatrix} * I \quad G_y = \begin{bmatrix} 0 & 1 & 0 \\ 0 & 0 & 0 \\ 0 & -1 & 0 \end{bmatrix} * I$$

$$G_{TL} = \begin{bmatrix} 1 & 0 & 0 \\ 0 & 0 & 0 \\ 0 & 0 & -1 \end{bmatrix} * I \quad G_{TR} = \begin{bmatrix} 0 & 0 & 1 \\ 0 & 0 & 0 \\ -1 & 0 & 0 \end{bmatrix} * I$$

(6)

$$G = \sqrt{G_x^2 + G_y^2 + G_{TL}^2 + G_{TR}^2}$$

(7)

The resulting features from the four directional operators are combined to generate a relatively comprehensive edge feature output. The features extracted from all four convolutional kernels are then summed to obtain a complete set of features. After integration and extraction via the ReLu activation function and 3×3 convolutional kernels, the residual structure further emphasizes the features that need to be learned.

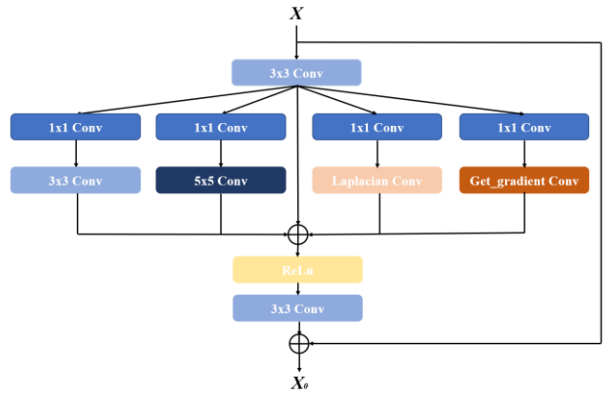

Fig3 Multiple convolution operator feature extraction modules(MSC)

## 3.2 Multiple sets of cross-feature extraction modules

Equation 8 describes a process for computing the output of a convolutional neural network (CNN) layer. The input to this layer is composed of two groups of features, denoted by $X_0$ and $X_1$. The first group of features $X_0$ is first passed through a softmax activation function, which produces a probability distribution over the possible values of $X_0$. This distribution is then multiplied element-wise with the second group of features $X_1$, resulting in a new set of features denoted by $X_1'$. The second group of features $X_1'$ is then passed through another softmax function, and the resulting probabilities are again multiplied with another group of features, called $X_2$. This process of soft-maxing and element-wise multiplication is repeated until a new set of features $X_3'$ is obtained. At this point, the feature $X_3'$ is repeated across all channels of a 3x3 convolutional filter, producing a set of output features that are convolved with the initial features $X_0$. Finally, the result of this convolution is combined with the residual structure, which is a standard practice in residual networks. In summary, this equation describes a multi-stage process for extracting features from the input image that involves multiple softmax operations and cross-channel interactions, followed by convolution and residual connection.

$$\begin{aligned} X_0、X_1、X_2、X_3 &= DConv(X) \\ X_1' &= X_1 * \operatorname{soft max}(X_0) \\ X_2' &= X_2 * \operatorname{soft max}(X_1') \\ X_3' &= X_3 * \operatorname{soft max}(X_2') \\ Y &= Conv(Cat(X_0,\ X_1',\ X_2',\ X_3')) + X \end{aligned}$$

(8)

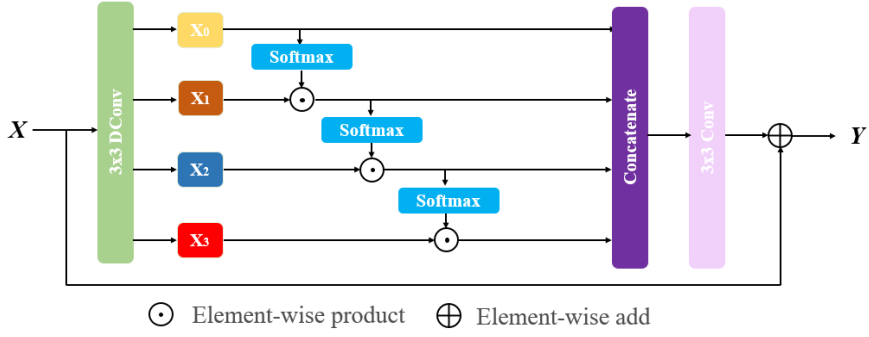

Fig4　Multiple sets of cross-feature extraction modules(MCO)

**3.3 Attention-based sparse graph neural network module**

The input features X in the module, where h×w represents the feature size and c represents the number of channels. The node feature matrix H can be obtained by reshaping the features, and the node feature matrix size is N × c, where N = h × w. The adjacency matrix *A* encodes the relationship between the nodes, and the graph *G* = (*A*, *H*) is constructed by successfully mapping each feature point in the feature to the corresponding graph node through X and A. The graph *G* = (*A*, *H*) is constructed through G The relationship between feature points is established to provide accurate feature information for global inference. The pixel points are mapped into the feature points, and the relationship between the pixel nodes is constructed through the equation shown in Equation 9. In the equation to ensure that the diagonal elements of S are zero, I in the equation is a unit matrix of size N*N, which represents the Hadamard product operation. The obtained feature similarity matrix *S* can then be regarded as a densely connected adjacency matrix. Then, we construct the graph Laplacian matrix by computing the normalized node connectivity distribution matrix *S*, where *D* is the degree matrix of *S*.

$$S = HH^T - HH^T \odot I$$
$$\overline{S} = D^{\frac{1}{2}} S D^{-\frac{1}{2}}$$
(9)

We believe that it is not necessary to compute all the relationships between pixel feature points using graph neural networks only the relevant features are required to compute them, and to achieve this goal, a Sparse Connection Adjacency Matrix needs to be constructed. By using Spherical Locality Sensitive Hashing (LSH) in NLSN[53] to form Attention buckets, features can be grouped by their angular distance. If the angles of two feature vectors are small, they are likely to be grouped in the same group. The features are sorted in descending order by the number of features in the group, and the number of calculated features is determined by choosing the parameter α. For example, if the parameter α is 0.5, the top 50% of the parameters that have been sorted are calculated, and A is modified according to the parameters that need to be calculated, and the associations of non-calculated features are cropped, and finally the Sparse Connection Adjacency Matrix $\overline{A}$ is obtained . The final construction of the $G_{sparse}$.

$$\overline{A} = \{\overline{S} \mid \overline{S}_{ij} \in NLSN(top[N_{num}*\alpha])\}$$
(10)

$$G_{sparse} = (\overline{A}, H)$$
(11)

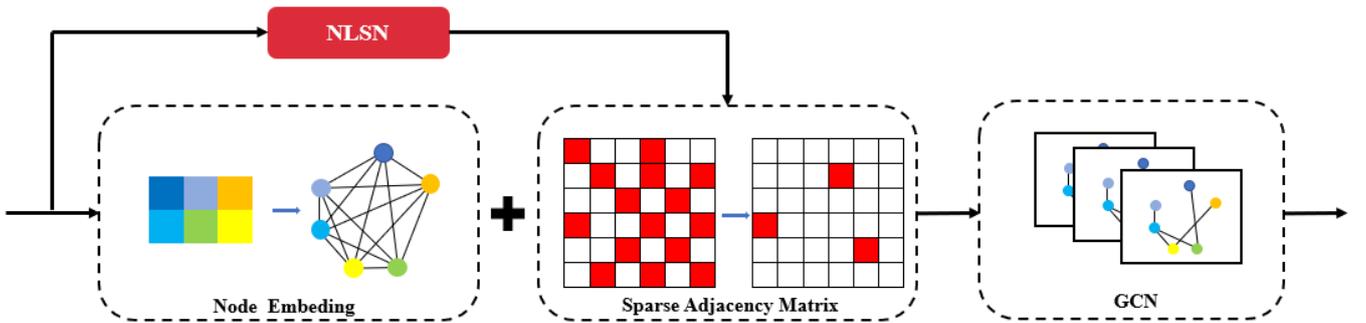

Fig5　Attention-based sparse graph neural network module(ASGNN)

**4. Experiments**

This section describes the evaluation of the network's performance through several experimentation studies. Initially, the details of the experimental setups are presented, followed by ablation experiments to assess the impact of key network components. Next, the network's performance was benchmarked against existing state-of-the-art image Super Resolution algorithms, and finally, the results were analyzed and summarized.

**4.1 Datasets**

Two sets of MRI data were selected for testing and training in order to verify the validity of the model.One set of MRI data was chosen in the Gamma Knife MR/CT/RTSTRUCT Sets With Hippocampal Contours (GammaKnife-Hippocampal) from the Cancer Imaging Archive[57]. Among them, 416 MRI images were randomly selected as the training set and 106 as the test set, all with image size of 512x512. Another set of data was selected from IXI Dataset(https://brain-development.org/ixi-dataset/), 1507 images were randomly selected as the training set and 376 images were selected as the test set, all of which were 256x256.The data images used in the training model all use Bicubic interpolation to conduct Downsampling to obtain images of the size required for training. The Downsampling images are processed by Gaussian blur, where the size of the Gaussian kernel is 7, and the noise with a standard deviation of 0.01 is added to the image.

**4.2 Experimental setup**

All models in this paper are implemented by pytorch, using Windows system, Intel(R) Xeon(R) Gold6240R CPU, TeslaA100 40G GPU .Both models are trained using the AdamW optimizer with $\beta_1 = 0.99$, $\beta_2 = 0.999$, weight_decay=0.01 and $\varepsilon = 10^{-8}$ for 50 epochs. Batch size is chosen as 6. The learning rate is 0.0001. Due to the good convergence of the L1 pixel loss function[37], we choose it as the loss function for network training as a way to illustrate the effectiveness of our proposed network.

**4.3 Evaluation metrics**

The performance of our proposed method has been systematically evaluated by comparing it with various established image super-resolution (SR) methods. Two popular evaluation measures have been employed, namely, Peak Signal-To-Noise Ratio (PSNR) and Structural Similarity Index Measure (SSIM). PSNR serves as a commonly utilised metric in evaluating the image quality, measuring the difference between original images and their respective reconstructions. With units expressed in decibels, a higher PSNR score indicates better performance. On the other hand, SSIM captures the degree of resemblance between images. A nearer the value of 1 implies a greater similarity between the images being assessed. By combining these two metrics, we can offer an overall perspective of the efficacy of the developed approach when contrasted with others.**4.4 Rseult**

To verify the reconstruction performance of our proposed model, we compare it with the state-of-the-art classical SR model. The classical SR models used for comparison include Cross_srn [39], DeFiAN [40], RFDN [41], LCRCA [42], LESRCNN [43], SwimIR [44], VDSR [45], SRCNN [46], FSRCNN [47], T2Net [48], VapSR [49] , Divanet [50], ESRT [51], and LBNet [52]. Using the above models and our proposed models, the datasets GKH and IXI with different degrees of degradation were recovered by image reconstruction and the PSNR and SSIM values were obtained by comparing the recovered models with the actual images. The PSNR and SSIM values obtained for each model are shown in Table 1.

Table1 . Quantitative comparison with the state-of-the-art method on a benchmark dataset. The top two results are shown in black and blue, respectively.

| Methods | Scale | GKH | IXI | Scale | GKH | IXI |
|---|---|---|---|---|---|---|
| Bicubic | ×2 | 34.7750/0.9154 | 26.6179/0.7699 | ×4 | 32.4887/0.8795 | 25.1573/0.6908 |
| cross_srn | ×2 | 41.7520/0.9730 | 31.6307/0.9207 | ×4 | 37.6165/0.9505 | 27.7552/0.8312 |
| defian | ×2 | *41.7809*/0.9736 | *31.6878*/0.9212 | ×4 | 37.7397/0.9521 | 28.1233/0.8426 |
| RFDN | ×2 | 41.6938/0.9735 | 31.5964/0.9211 | ×4 | 37.5444/0.9508 | 28.2015/0.8461 |
| grnn | ×2 | 41.6965/0.9730 | 31.3143/0.9154 | ×4 | 37.6480/0.9502 | 27.8475/0.8338 |
| lesrcnn | ×2 | 41.4000/0.9721 | 31.4143/0.9176 | ×4 | 37.7408/0.9513 | 28.2170/0.8463 |
| swimIR | ×2 | 41.7014/0.9729 | 31.5725/0.9198 | ×4 | 37.8624/0.9526 | *28.3592/0.8523* |
| VDSR | ×2 | 40.6141/0.9548 | 31.3542/0.9081 | ×4 | 37.2706/0.9276 | 28.0758/0.8301 |
| SRCNN | ×2 | 40.9528/0.9701 | 30.7967/0.9074 | ×4 | 36.9578/0.9449 | 27.3907/0.8201 |
| FSRCNN | ×2 | 40.4784/0.9684 | 30.2108/0.8969 | ×4 | 36.1402/0.9393 | 26.1833/0.7712 |
| T2Net | ×2 | 41.7651/0.9735 | 31.4422/0.9171 | ×4 | 37.6930/0.9513 | 27.9939/0.8387 |
| VapSR | ×2 | 41.8051/*0.9741* | 31.6864/*0.9221* | ×4 | 37.6150/0.9532 | 28.0323/0.8421 |
| Divanet | ×2 | 41.7403/0.9736 | 31.2796/0.9152 | ×4 | *37.8637*/**0.9537** | 28.2351/0.8448 |
| LBNet | ×2 | 40.3056/0.9649 | 31.0625/0.9053 | ×4 | 36.4150/0.9375 | 27.6641/0.8263 |
| ESRT | ×2 | 41.5532/0.9731 | 31.5515/0.9199 | ×4 | 37.6147/0.9519 | 27.9254/0.8375 |
| **Ours** | ×2 | **41.9420/0.9746** | **31.8535/0.9246** | ×4 | **37.9132**/*0.9535* | **28.4816/0.8531** |

Based on experimental comparisons of different models for super-resolution reconstruction on various datasets with differing image sizes, our proposed model demonstrated higher PSNR and SSIM scores than competitor models. Specifically, in the GKH dataset, for X2 and X4 magnified images, our proposed model outperformed the second-place model by 0.16 dB and 0.05 dB in terms of PSNR, respectively, and had comparable SSIM scores. Similarly, in the IXI dataset, our proposed model outperformed the second-place model by 0.17 dB and 0.13 dB, respectively, in terms of PSNR for X2 and X4 magnified images. Overall, our proposed model demonstrated higher PSNR and SSIM scores across all datasets, thus confirming its superiority over other models.

To further understand the super-resolution reconstruction capability of our proposed model, images formed by cross_srn, defian, Divanet, ESRT, LBNet, swimIR, and VapSR models were selected for comparison. The images in Figure 6 after super-resolution reconstruction of multiple models with different degradation levels in different datasets. We can find that the recovered images of cross_srnd, defian, and Divanet have insufficient details and blurred boundaries, and the reconstructed images of ESRT, LBNet, swimIR, and VapSR have intermittent problems, although they can be recovered with narrow and thin lines and other shapes.

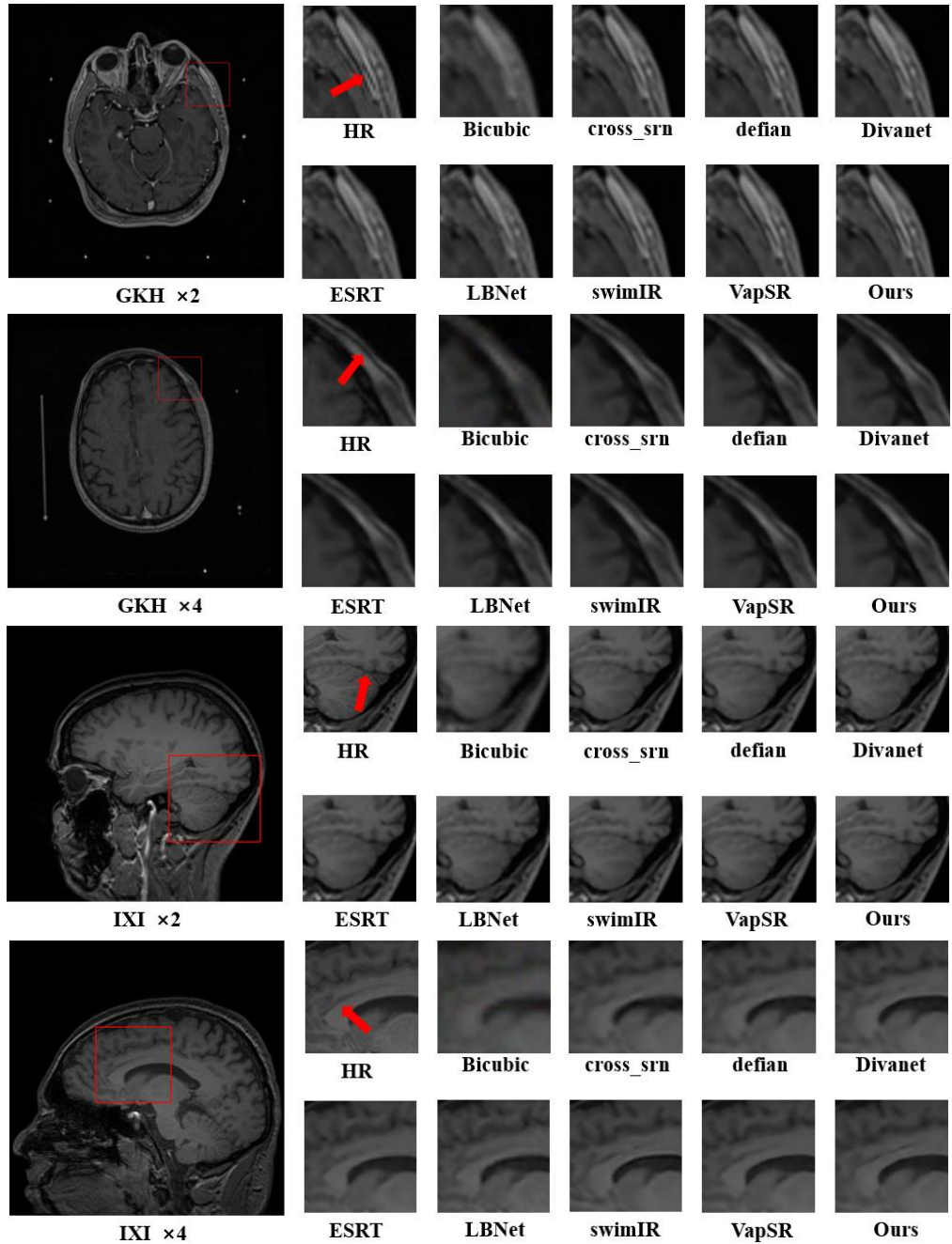

Fig6 Bicubic、cross_ SRN, Defense, Divanet, ESRT, LBNet, SwimIR, VapSR, and our proposed model have image reconstruction capabilities under different magnifications on multiple datasets.

## 4.5 Ablation Study and discussion

Multiple sets of ablation experiments were conducted to verify the image reconstruction ability of the model's modules and parameters for super-resolution images.

Table2 the reconstruction effect of different data images at different magnifications when using bicubic operation in the model or not

| Parameters | Scale | GKH | IXI | Scale | GKH | IXI |
|---|---|---|---|---|---|---|
| No BiCubic | ×2 | 41.5936/0.9738 | 31.8061/0.9244 | ×4 | 37.9125/0.9528 | 28.4181/0.8507 |
| BiCubic | ×2 | 41.9420/0.9746 | 31.8535/0.9246 | ×4 | 37.9132/0.9535 | 28.4816/0.8531 |

The results of the ablation experiments in the table show that the bicubic operation can effectively improve the super-resolution image reconstruction results, and the most obvious improvement is 0.3484 dB for the image reconstruction effect of GKH data after the image is enlarged by 2 times.

Table3 Effect of the number of FEMs on the effect of image super-resolution reconstruction

| Parameters | Scale | GKH | IXI | Scale | GKH | IXI |
|---|---|---|---|---|---|---|
| 3 | ×2 | 41.9420/0.9746 | 31.8535/0.9246 | ×4 | 37.9132/0.9535 | 28.4816/0.8531 |
| 4 | ×2 | 41.9289/0.9743 | 31.8475/0.9250 | ×4 | 37.9073/0.9537 | 28.4666/0.8522 |

By constructing FEMs with different numbers of blocks in the model, the effect of the number of FEMs on the image reconstruction effect is verified by Table 3 we can observe that the GHK and IXI data achieve better reconstruction effects at different magnifications when the number of FEMs is 3. The ability of the model to reconstruct images did not enhance with the increase in depth of the model.

Table4 Effect of ASGNN on image super-resolution reconstruction effect

| Parameters | Scale | GKH | IXI | Scale | GKH | IXI |
|---|---|---|---|---|---|---|
| No Attention | ×2 | 41.9095/0.9742 | 31.3352/0.9145 | ×4 | 37.9012/0.9543 | 28.3063/0.8473 |
| Attention | ×2 | 41.9420/0.9746 | 31.8535/0.9246 | ×4 | 37.9132/0.9535 | 28.4816/0.8531 |

To verify the effect of the ASGNN module on the model as a whole, ablation experiments were conducted by setting the same initial values of the model. It is obvious from Table 4 that adding the ASGNN module to the model can significantly improve the image reconstruction ability of the model especially in the IXI dataset, and the PSNR difference is 0.5183dB and 0.1753dB when the image is enlarged by 2X and 4X in the IXI dataset.

Table5 Calculation of the effect of parameter α on the effect of image super-resolution reconstruction

| Parameters | Scale | GKH | IXI | Scale | GKH | IXI |
|---|---|---|---|---|---|---|
| 0.5 | ×2 | 41.9325/0.9747 | 31.8363/0.9532 | ×4 | 37.8500/0.9532 | 28.5145/0.8538 |
| 0.75 | ×2 | 41.9420/0.9746 | 31.8535/0.9246 | ×4 | 37.9132/0.9535 | 28.4816/0.8531 |
| 1 | ×2 | 41.9517/0.9746 | 31.7836/0.9153 | ×4 | 37.8486/0.9536 | 28.4759/0.8475 |

To determine the image of the parameter α in the ASGNN module on the overall ability of the model, the same initial values of the model were set for the experiment. In the data of Table 5, it can be found that the overall image reconstruction effect of the model decreases when α = 1 compared with α = 0.5, while the overall image reconstruction capability of the model reaches the maximum when the value of α is 0.75.

Table6 Effect of different amounts of MCO and MSC on the super-resolution reconstruction of images

| MCO | MSC | Attention | Scale | GKH | IXI | Scale | GKH | IXI |
|---|---|---|---|---|---|---|---|---|
| 0 | 1 | √ | ×2 | 41.7292/0.9736 | 31.5357/0.9191 | ×4 | 37.8138/0.9524 | 28.1579/0.8436 |
| 1 | 0 | √ | ×2 | 41.9117/0.9744 | 31.8153/0.9237 | ×4 | 37.8932/0.9537 | 28.5039/0.8537 |
| 2 | 1 | √ | ×2 | 41.9478/0.9749 | 31.9599/0.9264 | ×4 | 37.8802/0.9535 | 28.5952/0.8569 |
| 1 | 2 | √ | ×2 | 41.9334/0.9747 | 31.8488/0.9246 | ×4 | 37.8997/0.9535 | 28.4680/0.8519 |
| 2 | 2 | √ | ×2 | 41.9372/0.9748 | 31.9509/0.9267 | ×4 | 37.8323/0.9538 | 28.5763/0.8564 |
| 2 | 2 | × | ×2 | 41.7580/0.9735 | 31.8499/0.9266 | ×4 | 37.9045/0.9542 | 28.5160/0.8539 |

To investigate the impact of different modules on model performance, we conducted experiments with various combinations of modules while keeping the model's initial parameters constant. Results from Table 6 revealed that as the number of MCOs increased while holding the number of MSCs constant, the model's ability to capture image features improved. However, this improvement gradually slowed down as the number of MCOs increased further. Conversely, when the number of MSCs was constant, the model's image capturing ability peaked when there was only one MCO, and it decreased when there were more.Interestingly, we observed that incorporating an ASGNN module into the model could effectively enhance the reconstruction recovery ability of the model, particularly when the number of MCOs and MSCs were equal. Overall, these findings demonstrate the importance of carefully considering module combinations in neural network design to achieve optimal model performance.

## 5. Conclusion

In this study, we propose a model for super-resolution reconstruction of MRI images. Our model includes multiple convolution operator feature extraction modules that use various convolution operators to capture image content and extract depth features. Additionally, multiple cross-feature extraction modules are deployed to further represent these extracted features. Since super-resolution image construction is sparse, we utilize an Attention-based Sparse Graph Neural Network module to speculatively fill the missing features by learning the relationships between blank and existing features and determining their relative impact on filling content. Our network outperforms state-of-the-art methods in both PSNR and SSIM when reconstructing clear images at different scales. Our future research aims to enhance model performance while expanding its applicability and reducing its size and complexity.

**Declaration of Competing Interest**

The authors declare that they have no known competing financial interests or personal relationships that could have appeared to influence the work reported in this paper.

**Acknowledgments**

This work was supported by the National Key Research and Development Program of China under Grant No. 2019YFB1311800, "Research on Minimally Invasive Cochlear Implant Robot System".